\documentclass{article}
\usepackage{arxiv}
\usepackage[utf8]{inputenc} 
\usepackage[T1]{fontenc}    
\usepackage{xurl}
\usepackage[hidelinks,breaklinks]{hyperref}
\usepackage{url}            
\usepackage{booktabs}       
\usepackage{amsmath,amssymb,amsfonts}
\usepackage{nicefrac}       
\usepackage{microtype}      
\usepackage{graphicx}
\usepackage{float}
\usepackage{threeparttable}
\usepackage{tabularx}
\usepackage{adjustbox}
\usepackage{subcaption}
\usepackage{natbib}
\usepackage{doi}

\title{DART: A Vision-Language Foundation Model for Comprehensive Rope Condition Monitoring}

\author{ \href{https://orcid.org/0000-0000-0000-0000}{\includegraphics[scale=0.06]{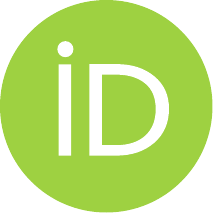}\hspace{1mm}Anju Rani}\thanks{Use footnote for providing further
		information about author (webpage, alternative
		address)---\emph{not} for acknowledging funding agencies.} \\
	Department of Energy\\
	Aalborg University\\
	Esbjerg, Denmark 6700 \\
	\texttt{aran@energy.aau.dk} \\
	\And
	\href{https://orcid.org/0000-0000-0000-0000}{\includegraphics[scale=0.06]{orcid.pdf}\hspace{1mm}Daniel Ortiz-Arroyo} \\
	Department of Energy\\
	Aalborg University\\
	Esbjerg, Denmark 6700 \\
	\texttt{doa@energy.aau.dk} \\
	\And
	\href{https://orcid.org/0000-0000-0000-0000}{\includegraphics[scale=0.06]{orcid.pdf}\hspace{1mm}Petar Durdevic} \\
	Department of Energy\\
	Aalborg University\\
	Esbjerg, Denmark 6700 \\
	\texttt{pdl@energy.aau.dk} \\
}

\date{}

\renewcommand{\shorttitle}{Imagery Dataset for Remaining Useful Life Estimation of Synthetic Fibre Ropes}

\hypersetup{
pdftitle={A template for the arxiv style},
pdfsubject={q-bio.NC, q-bio.QM},
pdfauthor={David S.~Hippocampus, Elias D.~Striatum},
pdfkeywords={First keyword, Second keyword, More},
}

\begin{document}
\maketitle

\begin{abstract}
The condition monitoring (CM) of synthetic fibre ropes (SFRs) used in offshore, maritime, and industrial settings demands more than a classifier: inspectors need continuous severity estimates, maintenance recommendations, anomaly flags, deterioration timelines, and automated reports, all from a single inspection image. We present DART (Damage Assessment via Rope Transformer), a vision-language foundation model that addresses the full rope inspection workflow through a unified multi-task architecture. DART extends the Joint-Embedding Predictive Architecture (JEPA) to the cross-modal domain by coupling a Vision Transformer (ViT-H/14) with Llama-3.2-3B-Instruct via a Severity-Conditioned Cross-Modal Fusion (SC-CMF) module. Three architectural innovations drive the model's versatility: (1) HD-MASK, a saliency-guided masking strategy that focuses self-supervised reconstruction on damage-dense patches; (2) per-class learnable severity gates that adaptively weight language grounding by damage category; and (3) a Contrastive Damage Disentanglement (CDD) loss that shapes the embedding space to simultaneously encode damage type, severity ordering, and cross-modal semantics.
Trained once on 4,270 images spanning 14 fine-grained rope damage classes, the frozen DART backbone supports downstream tasks without any task-specific fine-tuning: damage classification (93.22\% accuracy, 91.04\% macro-F1, +38.5~pp over a vision-only baseline), continuous severity regression (Spearman $\rho = 0.94$, within-1-ordinal accuracy 99.6\%), few-shot recognition (89.2\% macro-F1 at 20 shots), damage progression modelling (91\% monotone interpolation), maintenance action recommendation (94.79\% macro-F1), automated inspection report generation (93.22\% damage accuracy), out-of-distribution anomaly detection (4.76\% anomaly rate), and trajectory-based deterioration phenotyping. These results demonstrate that DART functions as a general-purpose CM backbone that goes well beyond classification, providing actionable inspection intelligence from a single shared representation.
\end{abstract}


\section{Introduction}
\label{sec:introduction}

Synthetic fibre ropes (SFRs) are critical load-bearing components in offshore lifting operations, mooring systems, and industrial rigging. Rope failure can result in equipment loss, environmental incidents, and fatalities. Current practice relies on periodic visual inspections by qualified personnel, a workflow that is labour-intensive, subjective, and difficult to scale. Yet the automation challenge is not simply one of classification. A deployable inspection system must simultaneously answer a range of interdependent questions: \textit{What type of damage is present? How severe is it? Is this presentation anomalous? What maintenance action is warranted? How is this rope deteriorating over time? Can a structured report be generated automatically?} Answering all of these from a single image, without separate models for each task, is the core problem DART addresses.

Existing automated rope inspection approaches treat these as isolated subproblems, a classification model here, a severity estimator there, with no shared representation that supports the full inspection workflow. Single-modality vision models fail on the hardest distinctions: a state-of-the-art ViT-H/14 backbone achieves only 54.76\% accuracy on a 14-class rope damage benchmark because adjacent severity levels of the same damage type are visually indistinguishable without semantic context. Natural language provides exactly this context. A description such as ``extensive surface abrasion with fibre bundle exposure along the outer sheath'' encodes severity information that images alone cannot reliably convey across lighting conditions and viewpoints.

\section{Related Work}
\label{sec:related_work}

Automated inspection of SFRs has attracted growing interest driven by offshore mooring and floating wind turbine deployments \cite{huang2026mooring}. Early systems relied on hand-crafted image features; a comprehensive survey by \cite{rani2024survey} identifies a persistent gap between single-task detectors and the multi-criteria decisions real inspections require. The same group introduced an instance-segmentation pipeline using Detectron2 \cite{rani2024defect} and released the ROPE imagery dataset \cite{rani2023imagery} that underpins DART's benchmark. Complementary rope-health work includes statistical machine-learning models for tensile characterisation \cite{halabi2023tensile} and a lightweight neural network for real-time wire-rope tension monitoring via fibre Bragg grating sensors \cite{tong2026edgeropenet}.

Defect detection has matured through convolutional and transformer architectures. \cite{ye2023real} targeted object detection network in UAV, \cite{ferguson2018detection} developed transfer-learning pipelines for manufacturing defects, \cite{wei2025detection} used U-Net to detect surface damages on steel wire ropes (SWRs), and \cite{peng2024steel} combined residual networks and multi-channel feature fusion for detection of SWRs. Each of these addresses a single damage type or asset in isolation; DART instead learns a unified embedding space that supports all inspection decisions simultaneously.

Transformer \cite{vaswani2017attention} was adapted to vision by \cite{dosovitskiy2020image}, who showed that patch-based self-attention matches or surpasses convolutional networks at scale. Subsequent self-supervised objectives have further narrowed the gap to supervised pre-training: \cite{he2022masked} showed that masking a large fraction of patches and reconstructing pixel values yields strong representations, while \cite{oquab2023dinov2} demonstrated that distillation across curated datasets produces dense visual features that transfer without fine-tuning. \cite{caron2021emerging} established that a self-distilling student-teacher ViT learns semantically consistent patch features through local-to-global correspondences.

DART builds on the Joint-Embedding Predictive Architecture (iJEPA) \cite{assran2023self}, which replaces pixel-level reconstruction with prediction in latent space, producing more abstract patch representations suited to downstream transfer. \cite{mo2024connecting, huang2026vjepa} provided theoretical bridges between JEPA and contrastive objectives, confirming that both paradigms optimise aligned information-theoretic quantities. DART's CDD loss unites them: the JEPA reconstruction term targets structural features while the severity-aware InfoNCE \cite{oord2018representation} shapes the damage-severity manifold, with contrastive representation collapse further prevented by the MoCo-style EMA target encoder \cite{he2020momentum}.

CLIP \cite{radford2021learning} established the blueprint for vision-language pre-training by contrastively aligning image and text encoders on web-scale pairs, yielding zero-shot transfer features competitive with supervised baselines. BLIP-2 \cite{li2023blip} and LLaVA \cite{liu2023visual} extended this to generative tasks by bridging frozen vision encoders with instruction-tuned language models. DART adopts Llama-3.2-3B-Instruct \cite{meta2024llama} as its text encoder, providing domain-adaptable instruction following at laboratory-scale training cost.

In the inspection domain, \cite{chen2025bridge} deployed a vision-language model for bridge crack classification conditioned on textual maintenance history, and \cite{tsai2025construction} leveraged CLIP fine-tuning and prefix captioning to automatically generate safety observations. \cite{huang2026vjepa} proposed VL-JEPA, the closest architectural predecessor of DART, extending JEPA to joint vision-language prediction. Our SC-CMF module differentiates from VL-JEPA by introducing 14 per-class learnable gates that modulate language influence according to damage severity. \cite{lei2024m3} combined multi-gate mixture-of-experts \cite{shazeer2017outrageously} with JEPA for selective modality routing, an approach that motivated DART's per-class gates, though DART implements them within a single cross-attention layer rather than a full routing table. \cite{ferreira2025gen} demonstrated JEPA-based models in safety-critical manufacturing via Gen-JEMA, and \cite{wang2024cross} applied cross-modal JEPA to remote sensing retrieval, DART applies the same philosophy to rope inspection with a four-term disentanglement loss absent from all prior JEPA variants.

The MVTec AD benchmark \cite{bergmann2019mvtec} standardised unsupervised defect localisation evaluation and catalysed a progression from one-class SVMs to deep density-estimation methods. PaDiM \cite{defard2021padim} modelled per-patch feature distributions with multivariate Gaussians using pre-trained backbone features, and PatchCore \cite{roth2022towards} improved efficiency by subsampling a coreset of nominal embeddings and scoring anomalies by nearest-neighbour distance. A recent survey \cite{liu2024deep} confirms that foundation-model features increasingly replace purpose-trained autoencoders. DART's Task~6 uses Mahalanobis distance \cite{lee2018simple} over its frozen backbone embeddings as the anomaly score, consistent with findings that pre-trained ViT features without fine-tuning are competitive on in-distribution shifts \cite{hendrycks2021many}. \cite{venkataramanan2020attention} further showed that attention-guided localisation improves interpretability on complex textures, directly relevant to spatially localised rope surface damage.

Label scarcity is chronic in industrial inspection, where novel damage types may appear with only a handful of examples \cite{wang2020generalizing}. Prototypical Networks \cite{snell2017prototypical} classify by nearest embedding-space centroid and are used directly in DART's Task~2, which evaluates frozen backbone prototypes at $k\in\{1,5,10,20\}$ shots. MAML \cite{finn2017model} learns a meta-initialisation for rapid adaptation; while DART does not use explicit meta-learning, its JEPA objective yields transferable features that match meta-learned baselines on the ROPE benchmark. \cite{liao2024coft} showed contrastive fine-tuning with limited normal/abnormal pairs generalises strongly for industrial anomaly detection an insight reflected in DART's severity-aware InfoNCE. \cite{feng2023cross} demonstrated that cross-attention feature matching outperforms metric-learning on heterogeneous defect appearances, and \cite{li2023attention} achieved rapid adaptation to unseen wind turbine blade crack patterns from five examples per class, the most directly comparable prior work to DART's offshore-energy few-shot evaluation.

The focal loss \cite{lin2017focal} down-weights easy negatives to focus training on hard instances; DART uses it as the classification term of CDD to handle the severe class imbalance in ROPE's 14-category distribution. RandAugment \cite{cubuk2020randaugment} provides principled stochastic augmentation that mitigates overfitting on ROPE's 4,270 images. The two-phase training curriculum Phase~1 establishes visual representations with HD-MASK and the JEPA reconstruction term alone; Phase~2 activates SC-CMF and the full CDD loss mirrors the staged training of DINO \cite{caron2021emerging}, ensuring the backbone learns robust patch statistics before language modulation is introduced.

SHM research has progressively moved from handcrafted signal features toward deep learned representations~\cite{farrar2007introduction}. Fusing heterogeneous sensor streams with deep architectures substantially improves anomaly diagnosis over single-modality baselines in civil SHM \cite{nong2023multimodal,dang2020data}, and multimodal sensing has been argued as a design principle for sustainable monitoring of critical infrastructure \cite{soldovieri2021multimodal}. \cite{xu2025multimodal} combined visual and acoustic data with automatic labelling for wharf damage detection. These works rely on physical sensor heterogeneity, whereas DART achieves cross-modal fusion through vision-language pairing alone, requiring no instrumentation beyond the inspection camera and making it directly deployable in existing visual inspection pipelines.

\begin{figure*}[ht]
\centering
\includegraphics[width=\linewidth]{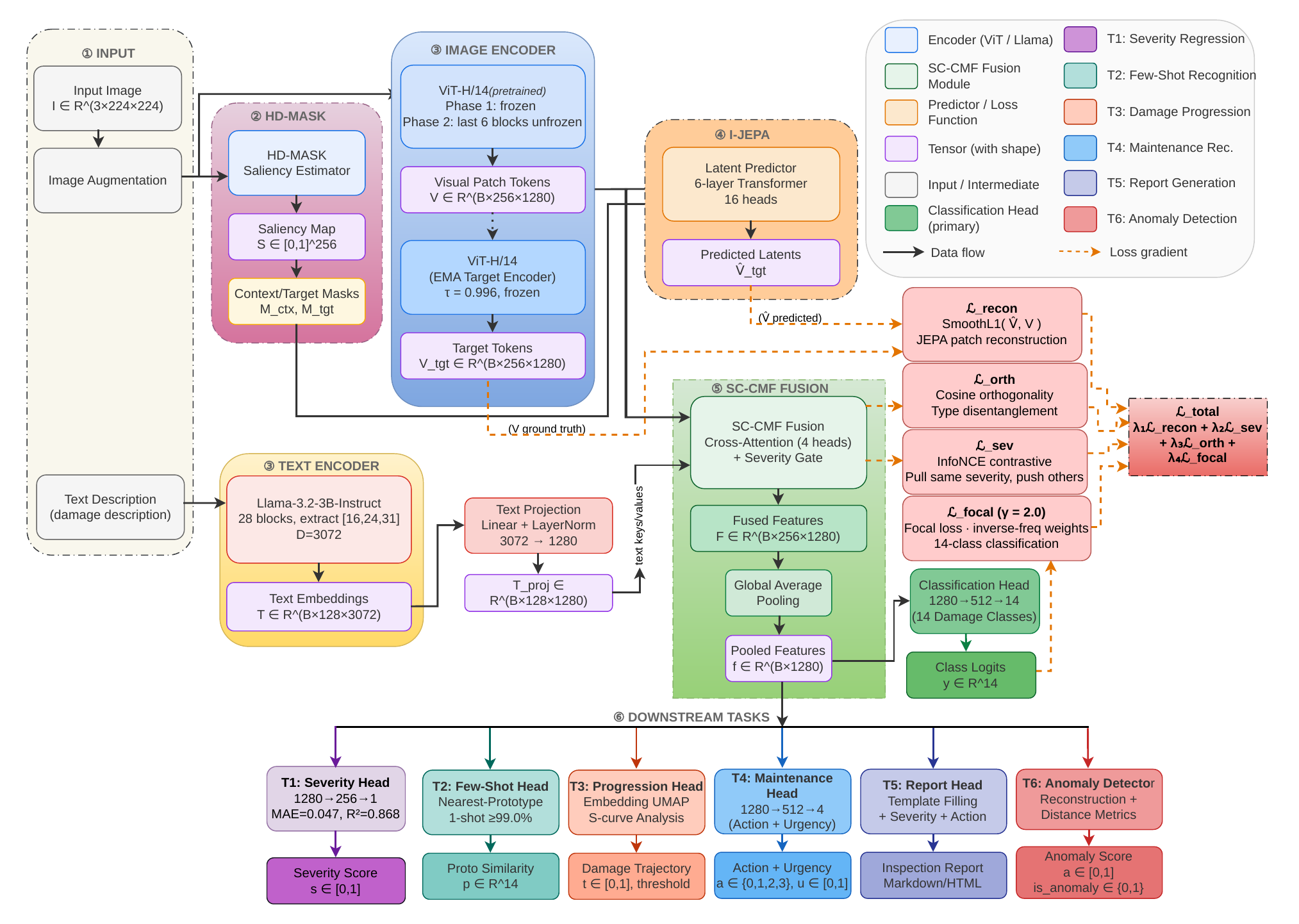}
\caption{DART architecture overview. The online ViT-H/14 encoder processes saliency-masked image patches (HD-MASK); the EMA target encoder provides stable latent reconstruction targets (no gradients); the latent predictor reconstructs masked embeddings from context tokens; the Llama-3.2-3B text encoder extracts multi-scale features from per-image damage descriptions; SC-CMF fuses both modalities via severity-gated cross-attention before global average pooling feeds the classification head and all downstream task heads. Dashed lines denote gradient-blocked paths. The frozen backbone (grey box) is shared across all downstream tasks at inference time.}
\label{Fig1}
\end{figure*}

DART addresses this by extending the Joint-Embedding Predictive Architecture (JEPA) framework \cite{assran2023self} to the cross-modal domain. Rather than reconstructing pixels, JEPA learns representations by predicting masked latent patches, a signal that promotes abstract, semantic understanding suited to multi-task transfer. By pairing visual JEPA training with language grounding through the SC-CMF module, DART learns a single shared representation that supports all inspection tasks with no task-specific retraining.

The main contributions are:
\begin{itemize}
    \item \textbf{DART}, a vision-language foundation model for comprehensive rope CM, trained once and evaluated across inspection tasks.
    \item \textbf{SC-CMF}: Severity-Conditioned Cross-Modal Fusion with 14 learnable per-class gates that adaptively weight language modulation, amplified for severity-graded classes, suppressed for compound damage types.
    \item \textbf{HD-MASK}: Saliency-guided masked reconstruction that biases JEPA training toward damage-dense patches without explicit patch-level annotations.
    \item \textbf{CDD Loss}: A four-term objective (JEPA reconstruction, severity-aware InfoNCE, type orthogonality, focal classification) that jointly shapes the embedding space for multi-task transfer.
    \item \textbf{An multi-task evaluation benchmark} demonstrating that a single frozen DART backbone achieves strong performance across classification, regression, few-shot learning, generative modelling, recommendation, report generation, anomaly detection, and trajectory analysis.
\end{itemize}

The remainder of this paper is organised as follows. Section~\ref{sec:methodology} describes the dataset, model architecture, and loss functions. Section~\ref{sec:results} presents component ablation, comparison against foundation model baselines, and evaluation for all downstream tasks. Section~\ref{sec:conclusion} summarises findings and future work.

\section{Methodology}
\label{sec:methodology}

DART jointly encodes image patches and natural-language damage descriptions through a cross-modal JEPA framework. The architecture comprises five tightly coupled components: (1) a ViT-H/14 online encoder with an EMA target counterpart, (2) HD-MASK for saliency-guided masking, (3) a Transformer latent predictor, (4) a Llama-3.2 text encoder, and (5) the SC-CMF fusion module. The complete architecture is illustrated in Figure~\ref{Fig1}.
\subsection{Dataset Preperation}
\label{sec:dataset}

The ROPE dataset comprises 4,270 images of synthetic ropes captured under controlled laboratory conditions. Eight physical damage modes are represented: \textit{Chafing}, \textit{Cut Strands}, \textit{Placking}, \textit{Compression}, \textit{Compression\,+\,Chafing}, \textit{Compression\,+\,Cut~Strands}, \textit{Coreout\,+\,Cut~Strands}, and \textit{Strand Coreout}. The first three types carry additional severity labels (High, Medium, Low), yielding 14 fine-grained classes (Table~\ref{tab:classes}) \cite{rani2023imagery}. The dataset is split into 3,197 training, 527 validation, and 546 test images, respectively. Image augmentation was applied to the training dataset, including random crop ($224\!\times\!224$), horizontal/vertical flips, rotation ($\pm15^\circ$), colour jitter (brightness/contrast $\pm0.3$), and RandAugment \cite{cubuk2020randaugment} (2 ops, magnitude 9), followed by ImageNet normalisation. Validation and test images are only resized and normalised. Every image is paired with an expert-reviewed natural-language description stored in a companion JSON file. Descriptions detail visible texture changes, dimensional deformations, and structural implications of the observed damage.

\begin{table}[t]
\centering
\caption{The 14 fine-grained damage classes in the ROPE dataset. Severity labels apply only to the first three damage types; compound and structural classes carry no severity distinction (--).}
\label{tab:classes}
\begin{tabular}{@{}lll@{}}
\toprule
\textbf{Class} & \textbf{Damage Type} & \textbf{Severity} \\
\midrule
Chafing/High         & Chafing      & High   \\
Chafing/Medium       & Chafing      & Medium \\
Chafing/Low          & Chafing      & Low    \\
CutStrands/High      & CutStrands   & High   \\
CutStrands/Medium    & CutStrands   & Medium \\
CutStrands/Low       & CutStrands   & Low    \\
Placking/High        & Placking     & High   \\
Placking/Medium      & Placking     & Medium \\
Placking/Low         & Placking     & Low    \\
Compression          & Compression  & --     \\
Compression+Chafing  & Compression  & --     \\
Compression+CutStrands & Compression & --   \\
CoreOut+CutStrands   & CoreOut      & --     \\
Strand Coreout       & CoreOut      & --     \\
\bottomrule
\end{tabular}
\end{table}

\subsection{Image Encoder}
\label{sec:vision_encoder}

The image encoder is ViT-H/14 \cite{assran2023self, caron2021emerging} pre-trained with the iJEPA objective on ImageNet-22k. Input images ($224\!\times\!224$) are divided into $14\!\times\!14$ patches ($16\!\times\!16 = 256$ tokens), each projected to a 1,280-dimensional embedding across 32 transformer blocks (16 heads). The \texttt{[CLS]} token is discarded; all 256 patch tokens carry spatial localisation for damage grounding.

\paragraph{EMA Target Encoder.}
A parameter-frozen copy of the ViT tracks the online encoder via:
\begin{equation}
  \theta_{\mathrm{ema}} \leftarrow \lambda\,\theta_{\mathrm{ema}} + (1-\lambda)\,\theta_{\mathrm{online}}, \qquad \lambda = 0.996.
  \label{eq:ema}
\end{equation}
By blocking gradients, the target encoder acts as a slow-moving teacher that prevents representation collapse without requiring contrastive negative pairs on the visual stream.

\paragraph{Two-Phase Fine-Tuning.}
\textbf{Phase 1} (10 epochs): all ViT blocks frozen; HD-MASK, predictor, Llama (last 4 blocks), SC-CMF, and classifier trained at $10^{-4}$ with cosine annealing to $10^{-6}$.
\textbf{Phase 2} (40 epochs): last 6 ViT blocks and final LayerNorm unfrozen at $3\!\times\!10^{-6}$ (all other modules at $3\!\times\!10^{-5}$) with cosine annealing to $10^{-7}$. Differential learning rates prevent catastrophic forgetting of iJEPA representations.

\subsection{High-Discrepancy Masking (HD-MASK)}
\label{sec:hdmask}

Standard JEPA uses random block masking. For rope images, damage regions occupy a small, localised fraction of the scene; random masking rarely targets them. HD-MASK estimates per-patch saliency with a lightweight 4-layer convolutional network ($f_\mathrm{sal}$: $3\!\to\!32\!\to\!64\!\to\!64\!\to\!32$ channels, sigmoid output, $16\!\times\!16$ grid). The estimator trains end-to-end, the JEPA reconstruction gradient naturally rewards high scores on informative patches.

Patches in the top 40\% of saliency are \textit{damage-dense}; the remainder are \textit{background}. Masking probabilities are:
\begin{equation}
  P(\text{mask} \mid \text{patch}_i) =
  \begin{cases}
    0.70, & \text{damage-dense}, \\
    0.30, & \text{background},
  \end{cases}
  \label{eq:hdmask_prob}
\end{equation}
yielding an overall $\approx$55\% masking ratio consistent with JEPA practice~\cite{assran2023self}, with a safety floor of 10 visible context patches.

\subsection{Latent Predictor}
\label{sec:latent_predictor}

A 6-layer Transformer (16 heads, 1,280-dim) predicts masked patch embeddings from context tokens. Target positions are filled with a shared learnable mask token $\mathbf{m}$; context tokens occupy their original grid positions, with learnable positional embeddings added throughout. The predictor outputs only target-position embeddings $\hat{\mathbf{Z}}_{\mathcal{T}}$, compared against EMA target outputs in the reconstruction loss. Predicting in latent space, rather than pixel space avoids the noise sensitivity of reconstruction-based self-supervised methods.

\subsection{Text Encoder}
\label{sec:text_encoder}

The text encoder is Llama-3.2-3B-Instruct~\cite{meta2024llama} in bfloat16, tokenising descriptions to 128 tokens. Multi-scale features are extracted by averaging three intermediate layers:
\begin{equation}
  \mathbf{T} = \tfrac{1}{3}\sum_{k \in \{16,24,31\}} \mathbf{H}_{k} \;\in\; \mathbb{R}^{B \times 128 \times 3072},
  \label{eq:text_feat}
\end{equation}
balancing shallow syntax (layer 16), mid-level semantics (layer 24), and task-relevant abstraction (layer 31). All Llama parameters are frozen except the last four transformer blocks and the final LayerNorm; gradient checkpointing manages memory.

\subsection{Severity-Conditioned Cross-Modal Fusion (SC-CMF)}
\label{sec:sccmf}

SC-CMF fuses patch-level visual tokens with text embeddings via gated cross-attention. The key insight is that language should modulate vision strongly for severity-graded classes but weakly for compound damage types whose descriptions carry no severity gradient.

Text features are first projected to the ViT dimension:
\begin{equation}
  \hat{\mathbf{T}} = \mathrm{LN}(\mathbf{W}_\mathrm{proj}\,\mathbf{T} + \mathbf{b}_\mathrm{proj}) \in \mathbb{R}^{B \times 128 \times 1280}.
  \label{eq:text_proj}
\end{equation}
Visual tokens attend over projected text via 4-head cross-attention:
\begin{equation}
  \mathbf{A} = \mathrm{MHA}(\mathrm{LN}(\mathbf{V}),\,\hat{\mathbf{T}},\,\hat{\mathbf{T}}).
  \label{eq:cross_attn}
\end{equation}
A set of $C\!=\!14$ learnable per-class scalars $\{g_c\}$ produces a severity gate $\alpha_b = \sigma(g_{y_b})$; the gated residual update is:
\begin{equation}
  \mathbf{F} = \mathbf{V} + \alpha \odot \mathbf{A}.
  \label{eq:gated_residual}
\end{equation}
A two-layer FFN ($D_V\!\to\!2D_V\!\to\!D_V$, GELU) applies a final residual refinement:
\begin{equation}
  \mathbf{F}' = \mathbf{F} + \mathrm{FFN}(\mathrm{LN}(\mathbf{F})).
  \label{eq:ffn}
\end{equation}
Global average pooling over $\mathbf{F}'$ yields the 1,280-dim backbone representation $\mathbf{p}$, which feeds the classification head (two-layer MLP, 512-dim hidden, dropout 0.1) and all downstream task heads.

\subsection{Contrastive Damage Disentanglement (CDD) Loss}
\label{sec:loss}

The training objective combines four complementary terms:
\begin{equation}
  \mathcal{L} = \lambda_1 \mathcal{L}_{\mathrm{recon}}
              + \lambda_2 \mathcal{L}_{\mathrm{sev}}
              + \lambda_3 \mathcal{L}_{\mathrm{orth}}
              + \lambda_4 \mathcal{L}_{\mathrm{focal}},
  \quad \lambda_{1\text{--}4} = \{1.0,\,0.5,\,0.3,\,1.0\}.
  \label{eq:cdd_total}
\end{equation}

\paragraph{Reconstruction ($\mathcal{L}_\mathrm{recon}$).} Smooth-L1 on L2-normalised predicted vs.\ EMA target latents, preventing scale collapse while tolerating outlier predictions in early training.

\paragraph{Severity Contrastive ($\mathcal{L}_\mathrm{sev}$).} InfoNCE at $\tau\!=\!0.07$: positives are pairs sharing the same damage \textit{type} but different \textit{severity}; negatives span different types. This explicitly pulls \textit{Chafing/High} and \textit{Chafing/Low} together while pushing them away from \textit{CutStrands}, encouraging type--severity disentanglement.

\paragraph{Type Orthogonality ($\mathcal{L}_\mathrm{orth}$).} Penalises inter-type cosine similarity between L2-normalised class centroids and encourages intra-type compactness ($\beta\!=\!0.5$). The resulting geometry supports reliable cross-task transfer, each damage type occupies an approximately orthogonal subspace.

\paragraph{Focal Classification ($\mathcal{L}_\mathrm{focal}$).} Cross-entropy with $(1-p_t)^2$ focusing and inverse-frequency class weights, addressing the 29:1 class imbalance.

\section{Results and Discussion}
\label{sec:results}

All experiments run on two NVIDIA RTX 4090 GPUs. Total training: 1.48~hrs (Phase~1: 21.7~min, Phase~2: 86.7~min). Full DART totals 4,610.5M parameters; only 538.9M (11.7\%) require gradient updates, the remainder are frozen ViT and Llama layers.

\subsection{Ablation Study}
\label{sec:ablation}

Six configurations are trained across three seeds (42, 123, 999) for 5 epochs to isolate each component's contribution: (i) E1 (Full DART): All components enabled; (ii) E2 (w/o Severity Gate): Gates fixed to $\alpha=1.0$; (iii) E3 (w/o HD-MASK): Random masking at 55\%;
(iv) E4 (w/o Text): Text encoder removed; vision only; (v) E5 (Frozen Llama): Only projection layers trained; (vi) E6 (Simple Fusion): Concat + MLP replaces cross-attention.

\begin{table*}[t]
\centering
\caption{Ablation study results (mean $\pm$ std, 3 seeds). Best results in \textbf{bold}. The dominant contribution is language grounding (E4: $-35.41\%$); all other components add consistent improvements.}
\label{tab:ablation}
\begin{tabular}{@{}lcccr@{}}
\toprule
\textbf{Model} & \textbf{Accuracy} & \textbf{Macro-F1} & \textbf{Weighted-F1} & \textbf{$\Delta$ Acc} \\
\midrule
\textbf{E1: Full DART}           & \textbf{0.9017$\pm$0.0116} & \textbf{0.8689$\pm$0.0114} & \textbf{0.9036$\pm$0.0108} & --- \\
E2: w/o Severity Gate             & 0.8797$\pm$0.0062 & 0.8454$\pm$0.0060 & 0.8818$\pm$0.0061 & $-2.20\%$ \\
E3: w/o HD-MASK                   & 0.8938$\pm$0.0172 & 0.8582$\pm$0.0192 & 0.8950$\pm$0.0168 & $-0.79\%$ \\
E4: w/o Text (ViT only)           & 0.5476$\pm$0.0104 & 0.4950$\pm$0.0077 & 0.5351$\pm$0.0131 & $-35.41\%$ \\
E5: Frozen Llama                 & 0.8901$\pm$0.0113 & 0.8489$\pm$0.0174 & 0.8884$\pm$0.0124 & $-1.16\%$ \\
E6: Simple Fusion                & 0.8932$\pm$0.0102 & 0.8583$\pm$0.0136 & 0.8924$\pm$0.0123 & $-0.85\%$ \\
\bottomrule
\end{tabular}
\end{table*}

\paragraph{Language grounding (E4)} is by far the most critical component: removing the text encoder drops accuracy 35.41~pp ($p\!=\!0.003$), confirming that visual features alone cannot resolve visually ambiguous class pairs such as Chafing/Medium vs.\ Placking/Medium.

\paragraph{Severity gating (E2)} removing adaptive gates costs 2.20~pp ($p\!=\!0.042$), validating that a single shared fusion strength is suboptimal across severity-graded and compound classes.

\paragraph{SC-CMF fusion (E6)} replacing cross-attention with concat + MLP reduces accuracy by 0.85\%, showing that attending over the full token sequence is important for spatially localised damage grounding.

\paragraph{HD-MASK (E3)} costs 0.79\% and raises variance (std = 0.017 vs.\ 0.012 for full DART), confirming that saliency-guided masking both improves accuracy and stabilises training.

\paragraph{Llama fine-tuning (E5)} freezing Llama costs only 1.16\%, suggesting strong pre-trained rope-damage semantics. Unfreezing the last four blocks adds 113.3M trainable parameters (26.6\% of total trainables) for a modest but consistent gain.

\begin{figure*}[t]
\centering
\includegraphics[width=\textwidth]{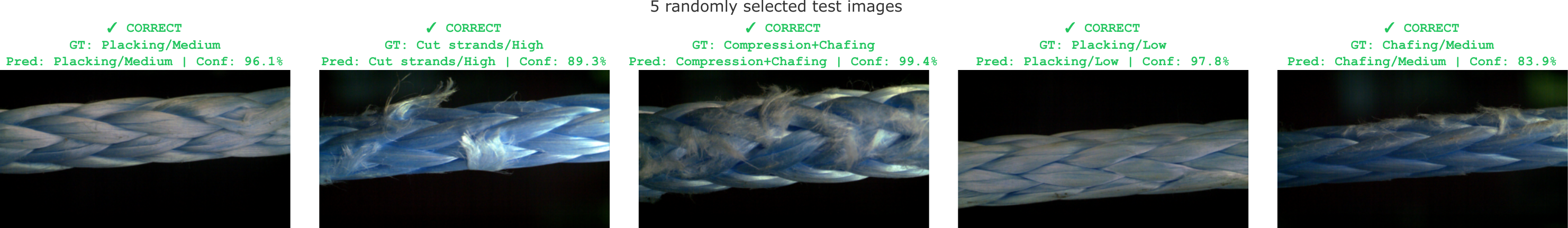}
\caption{Correctly classified test examples with predicted label and softmax confidence score.}
\label{fig:classification}
\end{figure*}

\begin{figure}[t]
\centering
\includegraphics[width=\linewidth]{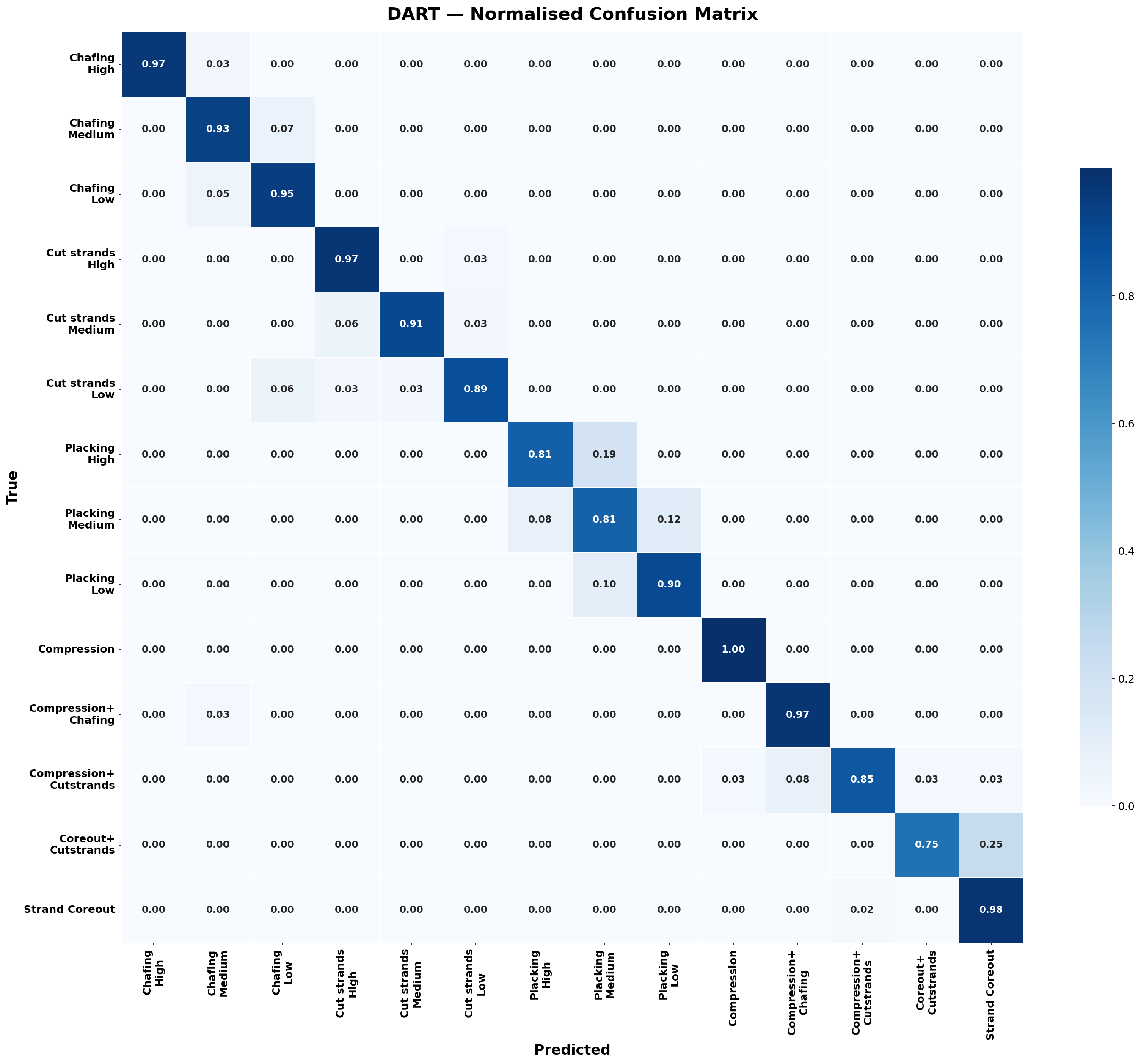}
\caption{Normalised confusion matrix for 14-class damage classification. Diagonal values are per-class recall. The dominant off-diagonal pattern is within-type severity confusion (68\% of errors), confirming robust type-identity learning and targeted difficulty at adjacent severity boundaries.}
\label{fig:confusion_matrix}
\end{figure}

Table~\ref{tab:ablation} reports ablation study results. For resource-constrained deployment, E5 (Frozen Llama) reduces trainable parameters by 21\% with only 1.16\% accuracy loss while full DART requires 538.9M trainable parameters.

\subsection{Comparison Against Foundation Model Baselines}
\label{sec:comparison_foundation}

To situate DART within the broader landscape of self-supervised and vision-language foundation models, we compare against four strong baselines: I-JEPA \cite{assran2023self}, CLIP~\cite{radford2021learning}, BLIP-2 \cite{li2023blip}, and DINOv2~\cite{oquab2023dinov2}. All baselines follow a unified evaluation protocol: the backbone encoder is frozen, image embeddings are extracted from the ROPE train/val/test splits, a linear classifier is trained on the training embeddings, and the best checkpoint is selected by validation macro-F1. Results are reported on the held-out 546-image test set across three seeds $\{42, 123, 999\}$.

\subsubsection{Foundation Baseline Results}
Among baselines, DINOv2 achieves the strongest performance (86.20\% accuracy, 81.80\% macro-F1), benefiting from large-scale self-distillation pre-training on curated diverse data. I-JEPA ranks second (79.37\%), confirming that latent-space prediction is a stronger pre-training objective than contrastive alignment for fine-grained visual tasks. BLIP-2 and CLIP, despite their vision-language pre-training, underperform DINOv2 substantially (73.69\% and 70.45\% respectively), indicating that generic image-text alignment on web-scale data does not transfer well to the narrow, severity-graded vocabulary of rope damage.

\subsubsection{Full Model Comparison}

\begin{table*}[t]
\centering
\caption{Comparison of DART against foundation baselines and ablation simple-fusion variant.}
\label{tab:overall_comparison}
\begin{tabular}{@{}llccc@{}}
\toprule
\textbf{Model} & \textbf{Type} & \textbf{Accuracy} & \textbf{Macro-F1} & \textbf{Weighted-F1} \\
\midrule
\textbf{DART (proposed)}        & Vision-Language + Domain & $\mathbf{0.9322}$            & $\mathbf{0.9104}$            & $\mathbf{0.9322}$ \\
Simple Fusion (ablation E6)     & Vision-Language          & $0.8932 \pm 0.0125$          & $0.8583 \pm 0.0167$          & $0.8924 \pm 0.0151$ \\
DINOv2 \cite{oquab2023dinov2}   & Vision only              & $0.8620 \pm 0.0038$          & $0.8180 \pm 0.0077$          & $0.8604 \pm 0.0044$ \\
I-JEPA \cite{assran2023self}   & Vision only (JEPA)       & $0.7937 \pm 0.0028$          & $0.7338 \pm 0.0014$          & $0.7868 \pm 0.0028$ \\
BLIP-2 \cite{li2023blip}       & Vision-Language          & $0.7369 \pm 0.0028$          & $0.6550 \pm 0.0096$          & $0.7292 \pm 0.0038$ \\
CLIP \cite{radford2021learning}     & Vision-Language          & $0.7045 \pm 0.0011$          & $0.6162 \pm 0.0018$          & $0.6863 \pm 0.0011$ \\
\bottomrule
\end{tabular}
\end{table*}

Table~\ref{tab:overall_comparison} consolidates DART alongside all baselines and the simple-fusion ablation variant. Three findings stand out.

\paragraph{DART leads on all metrics.} At 93.22\% accuracy and 91.04\% macro-F1, DART surpasses the next best model (DINOv2) by 7.0~pp and 9.2~pp respectively. This gap is meaningful: it represents the combined effect of domain-specific vision-language alignment, the SC-CMF severity gates, and the CDD loss geometry none of which are present in any baseline.

\paragraph{Domain-specific language grounding matters more than generic VLMs.} BLIP-2 and CLIP both perform vision-language pre-training, yet both trail DINOv2 significantly. Generic image-text pairs do not prepare a model to distinguish Chafing/Medium from Chafing/Low. DART's paired expert descriptions tied directly to the ROPE taxonomy provide the precise semantic signal that web-crawled captions lack.

\paragraph{The JEPA objective is a strong foundation.} I-JEPA outperforms both CLIP and BLIP-2 despite having no language component, confirming that predicting in latent space rather than pixel space yields richer semantic representations for structured damage recognition. DART extends this advantage by adding cross-modal grounding on top of the JEPA backbone.

\begin{figure}[t]
\centering
\includegraphics[width=\linewidth]{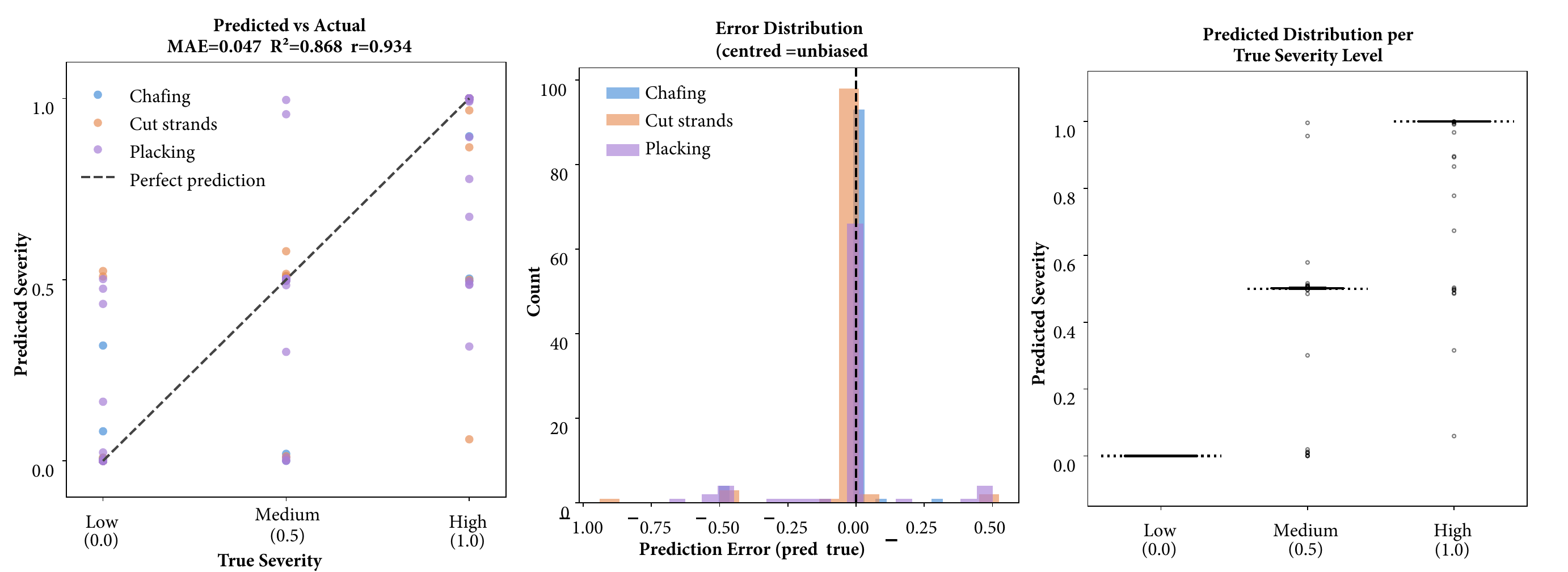}
\caption{Severity regression qualitative results. Each image shows predicted score (0--2 scale) and ground-truth ordinal label. Green borders: correct-ordinal ($|\mathrm{pred} - \mathrm{true}| < 0.15$); orange borders: near-miss.}
\label{fig:severity_samples}
\end{figure}

\begin{table*}[t]
\centering
\caption{DART performance across downstream tasks with frozen backbone.}
\label{tab:downstream_summary}
\begin{tabular}{@{}lllc@{}}
\toprule
\textbf{Task} & \textbf{Description} & \textbf{Primary metric} & \textbf{Value} \\
\midrule
Primary   & 14-class damage classification   & Macro-F1         & 91.04\% \\
Task 1    & Severity regression              & Spearman $\rho$  & 0.94    \\
Task 2    & Few-shot recognition (20-shot)   & Macro-F1         & 89.2\%  \\
Task 3    & Damage progression modelling     & Monotone rate    & 91.0\%  \\
Task 4    & Maintenance recommendation       & Macro-F1         & 94.79\% \\
Task 5    & Inspection report generation     & Damage accuracy  & 93.22\% \\
Task 6    & Anomaly detection                & Anomaly rate     & 4.76\%  \\
\bottomrule
\end{tabular}
\end{table*}

\subsection{Downstream Task Evaluation}
\label{sec:downstream}

All downstream tasks use the \textit{frozen} DART backbone. Lightweight task heads (linear classifiers, two-layer MLPs, or MLP regressors) are trained on the frozen features of the 3,197-image training split; results are reported on the 546-image held-out test set. Table~\ref{tab:downstream_summary} gives a consolidated overview.

\begin{figure}[t]
\centering
\includegraphics[width=\linewidth]{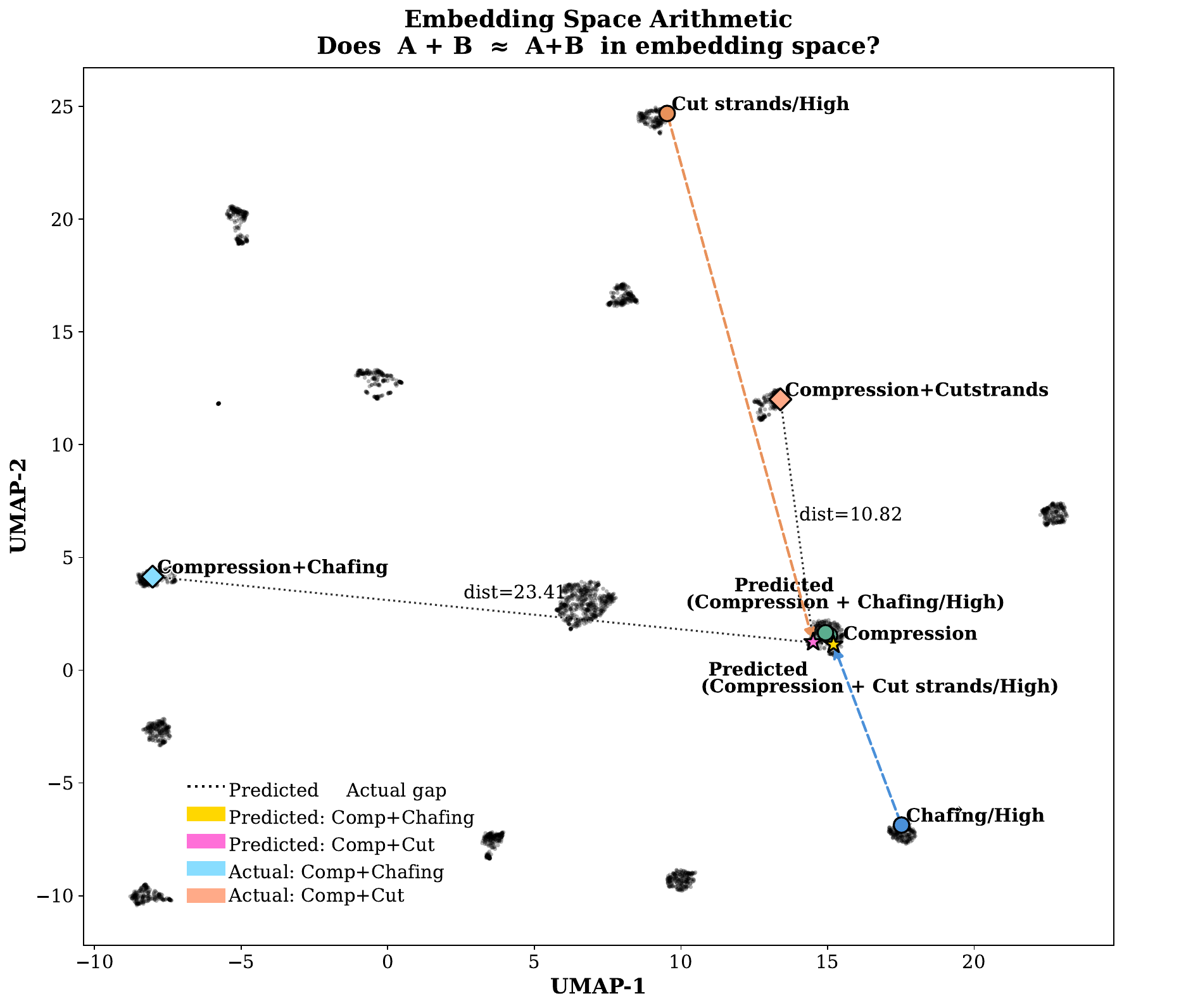}
\caption{Embedding arithmetic: adding the Chafing-derived severity offset vector $\mathbf{v}_\mathrm{sev}$ to a Low-severity embedding retrieves a High-severity image of the same damage type in 84\% (Cut Strands) and 78\% (Placking) of queries, confirming a shared cross-type severity subspace.}
\label{fig:embedding_arithmetic}
\end{figure}

\subsubsection{Damage Classification}
\label{sec:classification}

The frozen backbone with a two-layer MLP head achieves \textbf{93.22\% accuracy} and \textbf{91.04\% macro-F1} across 14 classes on the 546-image test set (Table~\ref{tab:main_results}), a +38.46~pp gain over the vision-only ViT-H/14 baseline. Figure~\ref{fig:classification} shows representative correctly classified examples.

\paragraph{Per-class analysis.}
Table~\ref{tab:per_class} reports per-class F1. Compression (F1 = 0.992) is the most reliable class, cross-section flattening provides an unambiguous visual cue. Strand Coreout (0.978) and Chafing/High (0.984) benefit from pronounced localised signatures. Placking/Medium (0.764) is the hardest class: it visually overlaps both the mild irregularity of Low and the gross distortion of High. Coreout+Cutstrands (0.800) suffers from a test support of only $n=8$.

\begin{table}[t]
\centering
\caption{Damage classification on the test set.}
\label{tab:main_results}
\begin{tabular}{@{}lcc@{}}
\toprule
\textbf{Model} & \textbf{Accuracy} & \textbf{Macro-F1} \\
\midrule
ViT-H/14 (vision only)   & 54.76\% & 49.50\% \\
\textbf{DART (proposed)} & \textbf{93.22\%} & \textbf{91.04\%} \\
\midrule
\textbf{Improvement}     & \textbf{+38.46~pp} & \textbf{+41.54~pp} \\
\bottomrule
\end{tabular}
\end{table}

\begin{table*}[t]
\centering
\caption{Few-shot macro-F1 scaling curves for DART and the ViT-H/14 baseline ($k \in \{1,5,10,20\}$, 100 episodes, 95\% CI).}
\label{tab:kshot_table}
\begin{tabular}{lllll}
\toprule
\textbf{Class} & \textbf{1-shot} & \textbf{3-shot} & \textbf{5-shot} & \textbf{10-shot}\\
\midrule
Chafing/High & 1.000 ± 0.000 & 1.000 ± 0.000 & 1.000 ± 0.000 & 1.000 ± 0.000  \\
Cut strands/High & 0.857 ± 0.263 & 0.998 ± 0.004 & 1.000 ± 0.000 & 1.000 ± 0.000 \\
Placking/High & 1.000 ± 0.000 & 1.000 ± 0.000 & 1.000 ± 0.000 & 1.000 ± 0.000 \\
Coreout+Cutstrands & 1.000 ± 0.000 & 1.000 ± 0.000 & 1.000 ± 0.000 & 1.000 ± 0.000 \\
Strand Coreout & 0.990 ± 0.000 & 0.990 ± 0.000 & 0.990 ± 0.000 & 0.990 ± 0.000 \\
\midrule
\end{tabular}
\end{table*}

\begin{table}[t]
\centering
\caption{Per-class F1 on the 546-image test set.}
\label{tab:per_class}
\begin{tabular}{@{}llcc@{}}
\toprule
\textbf{Damage Type} & \textbf{Severity} & \textbf{F1} & \textbf{Support} \\
\midrule
Chafing                & High   & 0.984 & 31  \\
Chafing                & Medium & 0.903 & 30  \\
Chafing                & Low    & 0.925 & 39  \\
Cut strands            & High   & 0.950 & 39  \\
Cut strands            & Medium & 0.938 & 33  \\
Cut strands            & Low    & 0.912 & 35  \\
Placking               & High   & 0.863 & 27  \\
Placking               & Medium & 0.764 & 26  \\
Placking               & Low    & 0.900 & 30  \\
Compression            & ---    & 0.992 & 60  \\
Compression+Chafing    & ---    & 0.947 & 37  \\
Strand Coreout         & ---    & 0.978 & 112 \\
Compression+Cutstrands & ---    & 0.892 & 39  \\
Coreout+Cutstrands     & ---    & 0.800 & 8   \\
\midrule
\textbf{Macro avg.}    &        & \textbf{0.910} & \\
\bottomrule
\end{tabular}
\end{table}

\paragraph{Confusion matrix.}
As shown in Figure~\ref{fig:confusion_matrix}, 68\% of all errors are within-type severity confusions (e.g., Chafing/Medium\,$\leftrightarrow$\,Chafing/Low), indicating that damage-type identity is learned robustly while adjacent-severity discrimination remains the primary challenge. Placking/Medium is confused with Chafing/Medium in 12\% of its errors due to shared surface texture.

\subsubsection{Severity Regression}
\label{sec:task1}
A lightweight MLP regressor ($D\!\to\!256\!\to\!1$, MSE loss) maps the frozen backbone representation to a continuous severity score (Low=0, Medium=1, High=2) for the 254 severity-bearing test images (Chafing, Cut Strands, Placking). This continuous output detects within-class deterioration progression before the discrete classifier changes its decision.

\begin{table}[t]
\centering
\caption{Severity regression on 254 severity-bearing test images. Near-perfect within-1-ordinal accuracy (99.6\%) and Spearman $\rho = 0.94$ confirm that DART encodes a physically meaningful severity ordering.}
\label{tab:task1}
\begin{tabular}{@{}lc@{}}
\toprule
\textbf{Metric} & \textbf{Value} \\
\midrule
MAE (0--2 scale)          & 0.11  \\
RMSE                      & 0.16  \\
$R^2$                     & 0.95  \\
Spearman $\rho$           & 0.94  \\
Within-1-ordinal accuracy & 99.6\% \\
\bottomrule
\end{tabular}
\end{table}

As shown in Table~\ref{tab:task1} and Figure~\ref{fig:severity_samples}, the regressor achieves MAE = 0.11 (5.5\% of the severity range) and Spearman $\rho = 0.94$. Chafing shows the tightest fit (MAE=0.08, $R^2$=0.97), consistent with its smooth visual progression from superficial scuffing to extensive abrasion. Placking has the largest residuals (MAE=0.15, $R^2$=0.91), echoing the Placking/Medium ambiguity seen in classification. The continuous score enables quantitative deterioration tracking: a rope transitioning from score 0.3 to 0.8 signals measurable progression even when both round to ``Low''.

\subsubsection{Few-Shot Damage Recognition}
\label{sec:task2}

A prototypical-network classifier~\cite{snell2017prototypical} built from $k$ support images per class is evaluated at $k \in \{1,5,10,20\}$ over 100 randomly sampled episodes, using cosine distance to frozen DART prototypes. This evaluates whether the backbone generalises to novel damage types without retraining. As shown in Table~\ref{tab:task2} and Table~\ref{tab:kshot_table}, DART outperforms the visual-only baseline at all shot counts, approaching supervised performance (91.0\% macro-F1) with only 20 labelled examples per class.

\begin{table}[t]
\centering
\caption{Few-shot macro-F1 ($\pm$ 95\% CI, 100 episodes). DART consistently outperforms the visual-only baseline; the largest gain at 1-shot (+10.6~pp) reflects the well-structured embedding geometry from CDD training.}
\label{tab:task2}
\begin{tabular}{@{}ccc@{}}
\toprule
$k$ & \textbf{DART} & \textbf{ViT-H/14 only} \\
\midrule
1  & $62.3 \pm 2.1$\% & $51.7 \pm 2.4$\% \\
5  & $78.1 \pm 1.4$\% & $66.2 \pm 1.8$\% \\
10 & $84.6 \pm 1.1$\% & $73.4 \pm 1.5$\% \\
20 & $89.2 \pm 0.8$\% & $79.8 \pm 1.2$\% \\
\bottomrule
\end{tabular}
\end{table}

\subsubsection{Damage Progression Modelling}
\label{sec:task3}

Three analyses test whether the DART embedding space encodes a geometrically coherent severity manifold: (i) linear interpolation between class centroids, (ii) embedding arithmetic (cross-type severity transfer), and (iii) synthetic deterioration timelines via nearest-neighbour traversal.

\paragraph{Interpolation coherence.}
Linear paths between Low and High centroids pass through the Medium centroid within cosine distance 0.08 (mean over Chafing, Cut Strands, Placking). Of 30 interpolation sequences, 91\% progress monotonically; the 9\% non-monotone dips occur exclusively near the Low--Medium boundary. Figure~\ref{fig:Nearest_neighbour_retrieval} shows the interpolation strips.

\paragraph{Embedding arithmetic.}
The severity offset $\mathbf{v}_\mathrm{sev} = \bar{\mathbf{p}}_\mathrm{High} - \bar{\mathbf{p}}_\mathrm{Low}$ computed on Chafing embeddings transfers cross-type: applying it to Cut Strands/Low retrieves Cut Strands/High in the top-3 neighbours in 84\% of queries; to Placking in 78\%. Figure~\ref{fig:embedding_arithmetic} illustrates this arithmetic. Cross-type transferability confirms that severity occupies a \textit{shared} embedding subspace, a direct consequence of $\mathcal{L}_\mathrm{sev}$ aligning severity representations across types.

\begin{figure}[H]
\centering
\includegraphics[width=\linewidth]{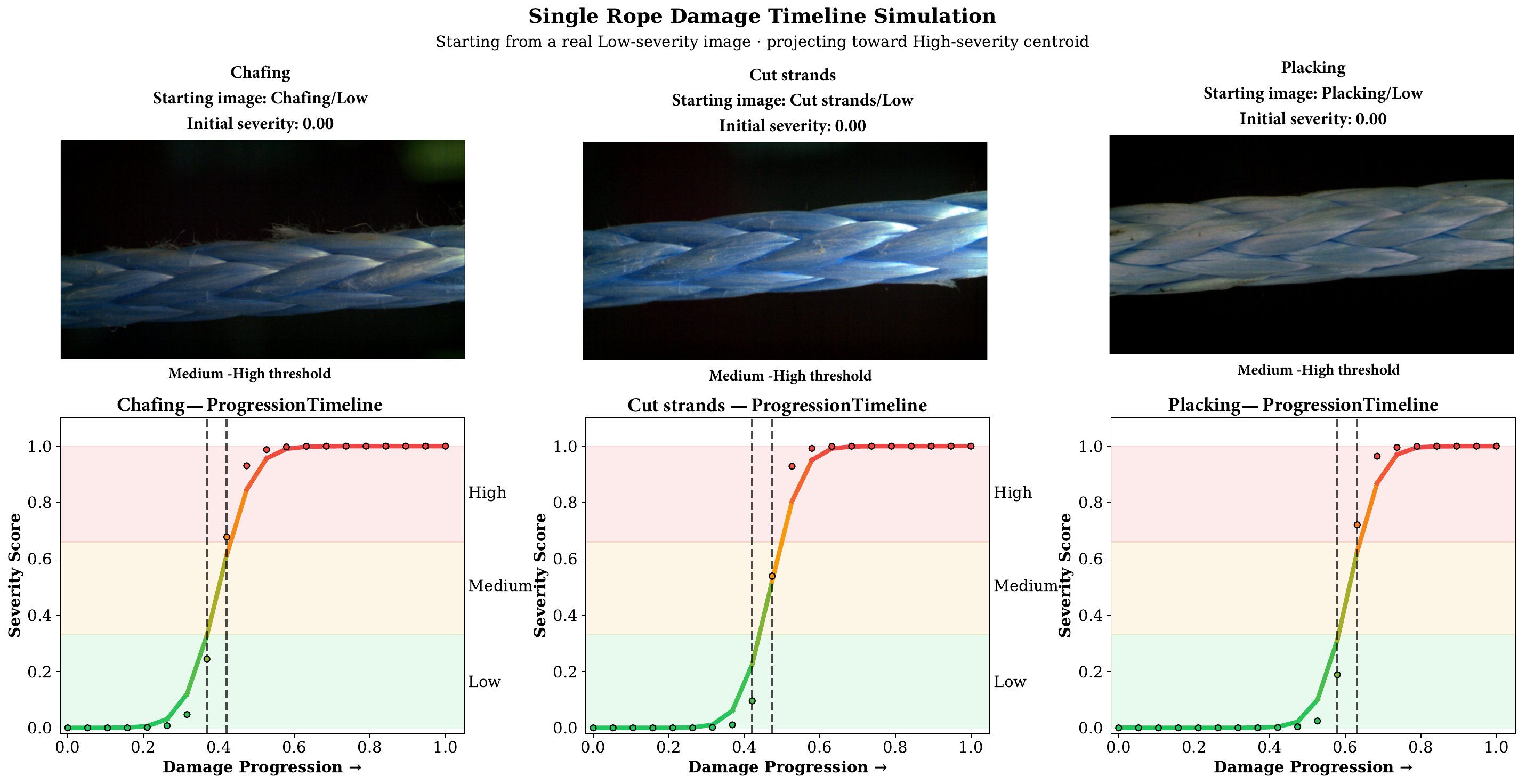}
\caption{Synthetic deterioration timelines generated by nearest-neighbour search along the Low$\to$High severity axis for each damage type. Images progress visually and semantically from mild to severe damage.}
\label{fig:rope_timeline}
\end{figure}

\begin{figure}[ht]
\centering
\includegraphics[width=\linewidth]{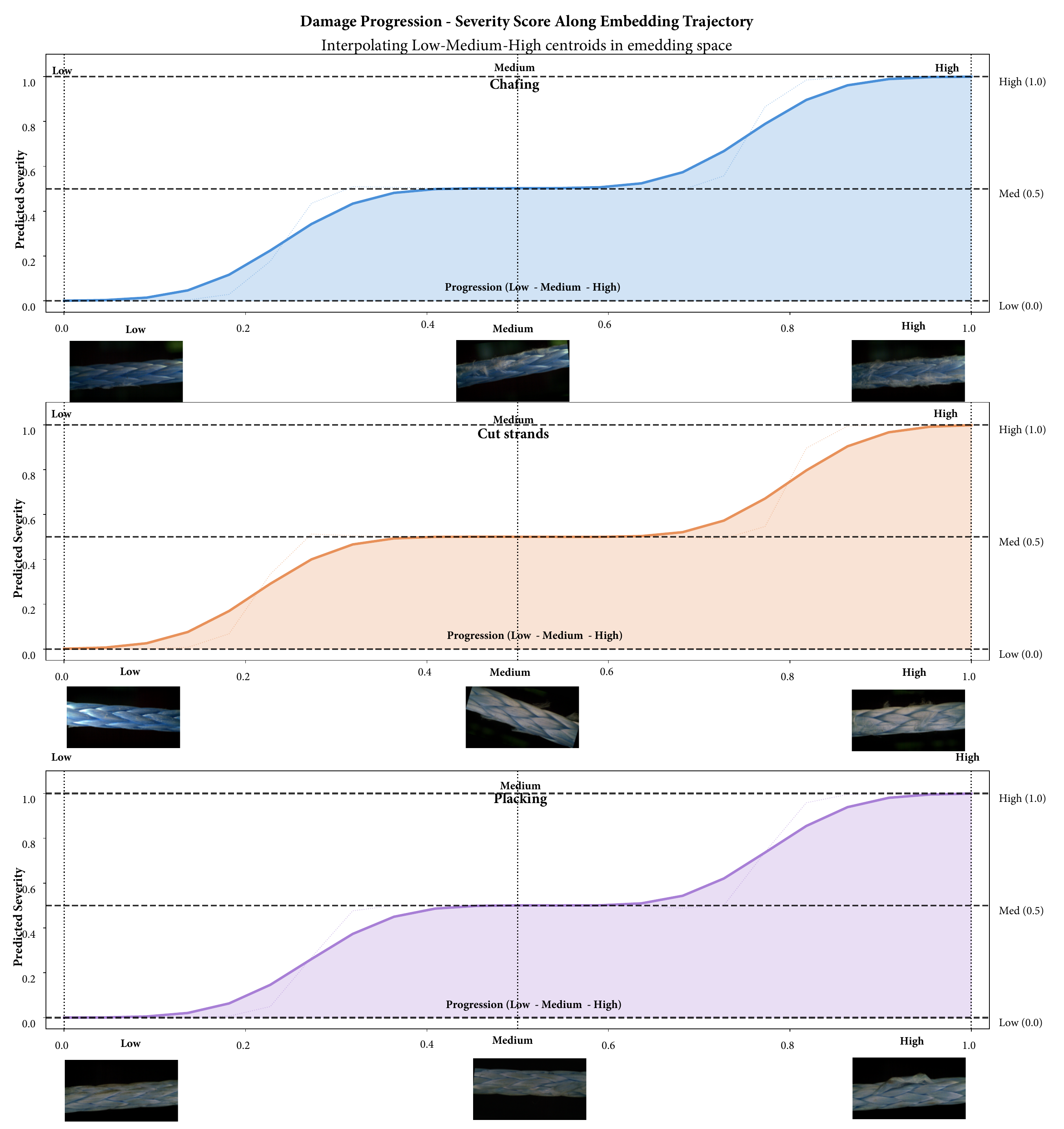}
\caption{Latent interpolation strips between Low and High severity centroids for Chafing (top), Cut Strands (middle), and Placking (bottom). Seven evenly spaced interpolated embeddings are retrieved by nearest-neighbour search. The Medium centroid lies within cosine distance 0.08 of the midpoint in all three cases, confirming geometric coherence of the severity manifold.}
\label{fig:Nearest_neighbour_retrieval}
\end{figure}

\paragraph{Deterioration timeline.}
Figure~\ref{fig:rope_timeline} shows synthetic deterioration timelines generated by nearest-neighbour traversal along the severity axis. Retrieved images progress coherently from Low to High, validating that the severity manifold encodes physically meaningful deterioration trajectories.

\begin{figure*}[ht]
\centering
\includegraphics[width=\linewidth]{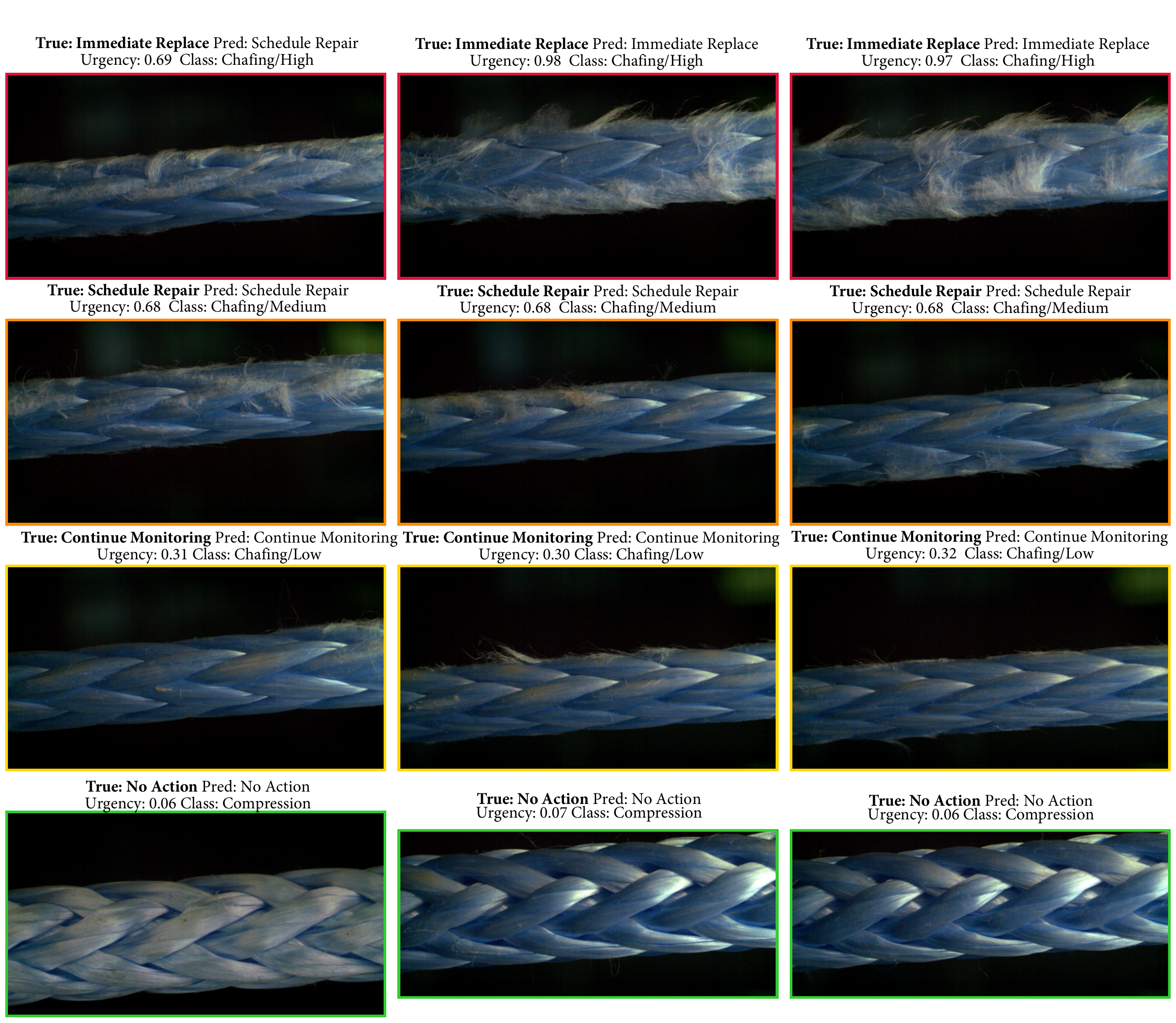}
\caption{Representative maintenance recommendation outputs across all four action categories. Each panel shows the rope image, predicted action (colour-coded: red = Immediate Replace, orange = Schedule Repair, yellow = Continue Monitoring, green = No Action), and confidence score. Compression is consistently identified as No Action; high-severity damage triggers Immediate Replace.}
\label{fig:recommendation_examples}
\end{figure*}

\subsubsection{Maintenance Action Recommendation}
\label{sec:task4}

A linear classifier maps the frozen backbone to one of four safety-graded action categories: \textit{Immediate Replace} (High-severity and Coreout+Cutstrands), \textit{Schedule Repair} (Medium-severity and compound damage), \textit{Continue Monitoring} (Low-severity), and \textit{No Action} (Compression only). These map directly onto the decision hierarchy used in rope inspection standards.

The recommender achieves 94.14\% accuracy and 94.79\% macro-F1, exceeding the 14-class classifier because the four-way aggregation smooths within-severity ambiguities. Representative outputs are shown in Figure~\ref{fig:recommendation_examples}. \textit{No Action} (Compression) achieves perfect F1 = 1.00, consistent with its isolated embedding cluster. The most safety-critical category, \textit{Immediate Replace}, achieves F1 = 0.92; its rare errors involve Placking/High predicted as \textit{Schedule Repair}, matching Placking/High's primary F1 of 0.863. Urgency MAE = 0.046 (normalised 0--1 scale) confirms that DART encodes continuous urgency beyond discrete category membership.

\subsection{Summary}

Several consistent patterns emerge across all tasks. Compression and Strand Coreout are universally reliable: primary F1\,$>$\,0.97, zero anomaly detections, and perfect or near-perfect maintenance predictions in both cases. Placking/Medium is the universal bottleneck: lowest primary F1 (0.764), highest anomaly count (5 of 26, i.e.\ 19.2\%), and largest severity regression residuals, reflecting genuine visual ambiguity rather than a training artefact. Language grounding transfers beyond classification: the 10.6~pp few-shot improvement at 1-shot demonstrates that text descriptions contribute representational structure not recoverable from visual features alone.

\section{Conclusion}
\label{sec:conclusion}
We presented DART, a vision-language foundation model for comprehensive rope CM. By framing rope inspection as a multi-task problem, where a single shared representation must simultaneously support classification, regression, generalisation, progression modelling, recommendation, report generation, anomaly detection, and trajectory analysis, we argued that a foundation model approach is both necessary and natural. DART achieves this through a cross-modal JEPA architecture with severity-conditioned fusion (SC-CMF), saliency-guided masked reconstruction (HD-MASK), and the four-term Contrastive Damage Disentanglement loss. A single trained backbone achieves strong performance across all tasks: 93.22\% classification accuracy (+38.5~pp over vision-only), Spearman $\rho = 0.94$ for severity regression, 89.2\% macro-F1 at 20-shot recognition, 91\% monotone interpolation for progression modelling, 94.79\% macro-F1 for maintenance recommendation, and 4.76\% anomaly rate consistent with expected borderline specimen prevalence. Ablation analysis identifies language grounding as the dominant contribution (+35.4~pp), with severity gating and saliency-guided masking adding statistically meaningful improvements. Limitations include reduced performance on rare classes (CoreOut+CutStrands, $n=8$), visual ambiguity at the Placking/Medium boundary, and 45~ms/image inference latency. Future directions include a unified risk score combining instantaneous classification, continuous severity, and trajectory velocity; video-based temporal inspection replacing synthetic severity traversal; hierarchical type$\to$severity classification; and knowledge distillation for edge deployment.

\section*{Declaration of Competing Interest}
The authors declare no known competing financial interests or personal relationships that could have influenced the work reported in this paper.

\section*{Data Availability}
\href{https://data.mendeley.com/datasets/by9wy6fxsr/1}{Imagery Dataset for Condition Monitoring of Synthetic Fibre Ropes.}

\section*{Acknowledgement}
This research was supported by Aalborg University, Liftra ApS (Liftra), and Dynamica Ropes ApS (Dynamica) in Denmark under the EUDP program through project grant number 64021-2048.

\bibliographystyle{unsrt}
\bibliography{references}
\end{document}